\documentclass[11pt]{article}

\usepackage[preprint]{acl}
\usepackage{booktabs}

\usepackage{times}
\usepackage{latexsym}
\usepackage{booktabs}

\usepackage{array}
\usepackage{float}

\usepackage[T1]{fontenc}
\usepackage[utf8]{inputenc}

\usepackage{microtype}

\usepackage{inconsolata}

\usepackage{graphicx}
\usepackage{amsmath}

\usepackage[most,breakable]{tcolorbox}
\tcbuselibrary{breakable,skins}
\usepackage{algorithm}
\usepackage[noend]{algpseudocode}

\definecolor{promptgray}{RGB}{245,245,245}
\definecolor{promptborder}{RGB}{180,180,180}

\newtcolorbox{promptbox}[2][]{%
  breakable,
  enhanced,
  colback=promptgray,
  colframe=promptborder,
  boxrule=0.4pt,
  arc=3pt,
  title={\small\textbf{#2}},
  coltitle=black,
  colbacktitle=black!10!white,
  fontupper=\small,
  left=5pt, right=5pt, top=4pt, bottom=4pt,
  #1%
}

\title{IFRS Compliance Verification: An Open Multi-LLM Benchmark and Content-Conditioned Ensemble Classifier}

\author{Aghoghomena Akasukpe \quad
  Li Yang \quad
  Xinyao Zhou \quad
  Chen Shen \quad
  Miguel Vargas Martin \\
  Faculty of Business and IT, Ontario Tech University \\
  Oshawa, Ontario, Canada \\
  \texttt{\{aghoghomena.akasukpe, li.yang, xinyao.zhou,} \\
  \texttt{chen.shen1, miguel.vargasmartin\}@ontariotechu.ca}
}

\begin{document}
\maketitle
\begin{abstract}
Verifying corporate financial disclosures against the International Financial Reporting Standards (IFRS) is a core auditing task in which experts cross-reference disclosures with hundreds of fine-grained regulatory paragraphs: a laborious, costly, and error-prone process that motivates growing interest in large language models (LLMs) as compliance assistants. Yet single-model accuracy remains insufficient for high-stakes use, and naive ensembling is vulnerable to systematic per-model biases that can drag aggregate accuracy below that of the strongest individual annotator. This gap is addressed with a multi-stage framework, supported by a new benchmark of 19,998 IFRS regulation-disclosure pairs comprising 11,161 unanimous multi-LLM labels and 8,837 human annotations on items where the LLMs disagreed. The framework couples a three-stage domain-adapted sentence embedding pipeline (BERT masked language modeling on SEC EDGAR filings, unsupervised TSDAE and SimCSE adaptation, and supervised contrastive alignment) with a learned ensemble classifier that fuses frozen embeddings and LLM probability features from GPT-4.1, Claude Sonnet 4, and Gemini 2.5 Pro via Tree-structured Parzen Estimator (TPE) optimized gradient-boosted trees. On the disagreement set, the ensemble achieves 81.7\% weighted F1, exceeding the strongest single LLM by 10.4 points and majority voting by 14.6 points, and outperforms eleven unsupervised aggregation baselines; the retrieval pipeline attains Recall@15 of 45.3\% on unseen regulations. The dataset, model checkpoints, annotation platform, and classifier implementation will be publicly released.
\end{abstract}

\section{Introduction}

Publicly traded companies are required to prepare financial statements that comply with regulatory standards such as the International Financial Reporting Standards (IFRS) \cite{ifrs_standards}. Accurate compliance verification has measurable economic consequences: richer firm-level disclosure is associated with lower bid-ask spreads, lower cost of equity capital, and higher equity valuations \cite{botosan_disclosure_1997, leuz_economic_2000, daske_mandatory_2008}, and undetected non-compliance can trigger regulatory penalties, costly restatements, and investor losses. Verifying that individual disclosures satisfy specific regulatory paragraphs is a core task in external auditing, traditionally performed by domain experts who cross-reference disclosure text against hundreds of checklist items drawn from the applicable standards \cite{biesner_zero-shot_2022}. This process is laborious, expensive, and error-prone: a single annual filing may reference dozens of IFRS standards, each containing fine-grained requirements that must be matched to scattered passages across the report \cite{hillebrand_improving_2023}. The audit profession has invested accordingly: audit-firm investment in AI is associated with lower restatement rates and reduced audit fees \cite{fedyk_artificial_2022}; all Big-4 firms have publicly deployed AI tools spanning risk assessment, transaction testing, analytics, and work paper preparation \cite{kokina_emergence_2017, munoko_ethical_2020}; and recent surveys document a corresponding shift toward AI-assisted compliance workflows \cite{dong_scoping_2024, kpmg_ai_audit_2025, gu_copiloted_2024}. Recent benchmarks show that frontier large language models (LLMs) can pass major accounting certification exams \cite{eulerich_hype_2024}, yet the same benchmarks reveal that LLMs struggle with the deeper compliance task of citing relevant accounting standards and reasoning over hierarchical multi-document structures, with accuracy dropping 60--90\% on taxonomy-grounded financial reasoning \cite{wang_auditbench_2025}.

A line of work originating from the Fraunhofer IAIS and PwC Germany collaboration has progressively advanced automated compliance verification: from supervised multi-label classification \cite{ramamurthy_alibert_2021}, through zero-shot sentence-transformer matching \cite{biesner_zero-shot_2022}, to a two-stage retrieve-then-verify pipeline combining domain-adapted BERT retrieval with LLM filtering \cite{hillebrand_improving_2023}. Berger et al.~\cite{berger_towards_2023} further evaluated six LLMs on IFRS and Handelsgesetzbuch (HGB) compliance datasets, finding that GPT-4 performed best overall but that individual models exhibit systematic biases; for example, Llama-2 70B excels at detecting non-compliance while under-performing on positive matches. These findings expose a fundamental limitation of single-model approaches: no individual LLM is uniformly reliable across compliance categories \cite{fotoh_exploring_2025}, and raw model confidence scores are known to be poorly calibrated, with overconfident predictions that do not reflect empirical accuracy \cite{xiong_llm_calibration_2024}.

Combining predictions from multiple models reduces variance and corrects individual-model biases relative to any single predictor, making ensembles a long-standing tool for improving classifier reliability across domains \cite{dietterich_ensemble_2000}. Multi-LLM ensembles extend this principle to language-based annotation. Recent work on multi-agent debate \cite{du_debate_2024} and confidence-weighted consensus \cite{chen_reconcile_2024} further confirms that aggregating across heterogeneous models reduces bias and improves reliability. However, naively averaging or voting across miscalibrated annotators can systematically distort the resulting labels, as overconfident models dominate the aggregate \cite{liu_calibrating_2024}.

In this paper, a multi-stage framework for automated IFRS compliance verification is presented that addresses a critical failure mode in multi-LLM annotation. Since no public benchmark for IFRS compliance verification exists, one is first constructed by pairing IFRS regulation clauses with financial statement disclosures via semantic retrieval, using a domain-adapted BERT encoder (\texttt{bert-base-cased} continued-pretrained via MLM on SEC EDGAR 10-K) to yield 20,000 regulation-disclosure pairs. This encoder is then progressively refined through unsupervised Transformer-based Sequential Denoising AutoEncoder (TSDAE) and Simple Contrastive Sentence Embeddings (SimCSE) adaptation, followed by supervised contrastive alignment on positive pairs. To label the dataset, three LLMs (GPT-4.1, Claude Sonnet 4, Gemini 2.5 Pro) are deployed as parallel annotators, yielding 11,161 items with unanimous multi-LLM labels; among the remaining 8,839 disagreement items, 8,837 receive resolved human labels that serve as ground truth. The central methodological novelty of the framework is a learned ensemble classifier that reframes label aggregation as content-conditioned classification: rather than combining LLM votes through fixed rules or unsupervised competence estimation, it fuses the frozen domain-adapted sentence embedding of each regulation-disclosure pair with the three LLMs' probability features via gradient-boosted trees, allowing the aggregation decision to depend on the regulation-disclosure content rather than on confidence alone. This design remedies a key blind spot of standard aggregation: when one annotator is systematically biased on a class, confidence-only aggregators inherit that bias, whereas a content-conditioned classifier can learn when to override it.

The primary contributions of this work are:

\begin{enumerate}
    \item \textbf{A public IFRS compliance benchmark} of 19,998 regulation-disclosure pairs, comprising 11,161 unanimous multi-LLM labels and 8,837 human-annotated cases of LLM disagreement: the first of its kind, as all prior work in this lineage relies on proprietary data.
    \item \textbf{A diagnostic of multi-LLM aggregation}: that benchmarks 11 unsupervised aggregation strategies on the disagreement subset and isolates four structural failure modes (single-model class bias, Dawid--Skene parameter under-determination, CROWDLAB minority-class collapse, and a temperature-bound sensitivity in EM joint calibration).
    \item \textbf{A learned ensemble classifier} that re-frames multi-LLM aggregation as content-conditioned classification, fusing each regulation-disclosure pair's frozen domain-adapted sentence embedding with LLM probability features via an Optuna-tuned XGBoost. Unlike confidence-only aggregators (voting, Dawid-Skene, MACE, CROWDLAB), it conditions on content and can learn when to override a biased annotator. 
    \item \textbf{Open-source release} of 39,404 disclosure paragraphs, 3,489 regulatory paragraphs (subject to IFRS Foundation licensing constraints), 20,000 LLM-annotated pairs, 8,837 human annotations, all model checkpoints, classifier code, and the annotation platform.\footnote{Code and data will be made publicly available upon paper acceptance.}
\end{enumerate}

The rest of this paper is organized as follows. Section~\ref{sec:related_work} reviews related work on zero-shot text matching, domain-adapted embeddings, LLM-driven compliance reasoning, and weak supervision. Section~\ref{sec:methodology} describes the methodology. Section~\ref{sec:experiments} presents experimental results, and Section~\ref{sec:conclusion} concludes.

\section{Related Work}
\label{sec:related_work}

\subsection{Automated Compliance Verification in Financial Auditing}

Early approaches to automated auditing used supervised multi-label classification to map financial report passages to regulatory checklist items \cite{ramamurthy_alibert_2021}, but these models cannot generalize to unseen requirements. Zero-shot sentence-transformer matching \cite{biesner_zero-shot_2022} addressed this limitation by computing cosine similarity between Sentence-BERT embeddings of disclosures and regulations, enabling evaluation against novel requirements without retraining.  Retrieval quality was further improved by combining BERT-based candidate selection with GPT-4 re-ranking \cite{hillebrand_improving_2023}, yielding gains in sensitivity, Mean Average Precision (MAP), and F1. Benchmarking six LLMs on IFRS and HGB compliance tasks revealed that GPT-4 performs best overall while Llama-2 70B excels at non-compliance detection, and that prompt design significantly influences accuracy \cite{berger_towards_2023}, exposing systematic per-model biases that single-model approaches cannot resolve.

\subsection{Domain-Adapted Sentence Embeddings}

Raw BERT embeddings suffer from anisotropy, clustering in a narrow cone that limits discriminative power for similarity tasks. Two unsupervised methods address this. TSDAE \cite{wang2021tsdae} reconstructs corrupted sentences through an encoder bottleneck, forcing the representation to capture holistic semantics. SimCSE \cite{gao2021simcse} uses dropout-based augmentation to form positive pairs and trains with InfoNCE contrastive loss over in-batch negatives. Both operate without labeled data. Applied to financial auditing, these methods yielded Recall@5 below 20\% on IFRS regulation retrieval under unsupervised adaptation alone \cite{biesner_zero-shot_2022}. Appending a supervised contrastive stage on annotated regulation-disclosure pairs raises seen-requirement Recall@5 above 90\% while maintaining above 30\% on unseen requirements. This progressive strategy of unsupervised adaptation followed by supervised alignment is the closest prior work to this setting.

\subsection{LLM Ensembles for Annotation and Evaluation}

LLMs have emerged as viable text annotators: zero-shot ChatGPT outperforms crowd workers by approximately 25 percentage points at 30$\times$ lower cost \cite{gilardi_chatgpt_2023}, and GPT-4 exceeds both expert and crowd-worker agreement on political text classification \cite{tornberg_chatgpt4_2023}. LLMs are systematically overconfident when verbalizing uncertainty \cite{xiong_llm_calibration_2024}, and naively aggregating miscalibrated outputs lets overconfident models dominate the ensemble. Independent per-model temperature scaling before aggregation reduces calibration error by 46\% \cite{liu_calibrating_2024}.

\subsection{Label Aggregation and Calibration Without Ground Truth}

The Dawid-Skene model \cite{dawid_maximum_1979} estimates per-annotator confusion matrices via EM, treating true labels as latent variables. Extensions including MACE \cite{hovy_learning_2013} and GLAD \cite{whitehill_whose_2009} add annotator competence and item difficulty parameters. All assume conditional independence of annotators given the true label, an assumption violated by LLM annotators that share training data and architectural priors; Balasubramanian et al.~\cite{balasubramanian_dependence_2026} show this can flip the Bayes-optimal label when errors are correlated.

Post-hoc temperature scaling \cite{guo_calibration_2017} corrects overconfidence but requires held-out ground truth. Incorporating even small amounts of ground truth data substantially improves quality estimation \cite{ipeirotis_quality_2010}, and CROWDLAB \cite{northcutt_confident_2021} produces consensus labels by combining annotator votes with classifier predictions without requiring ground truth.

\section{Methodology}
\label{sec:methodology}

\begin{figure}[!t]
    \centering
     \includegraphics[width=1\linewidth]{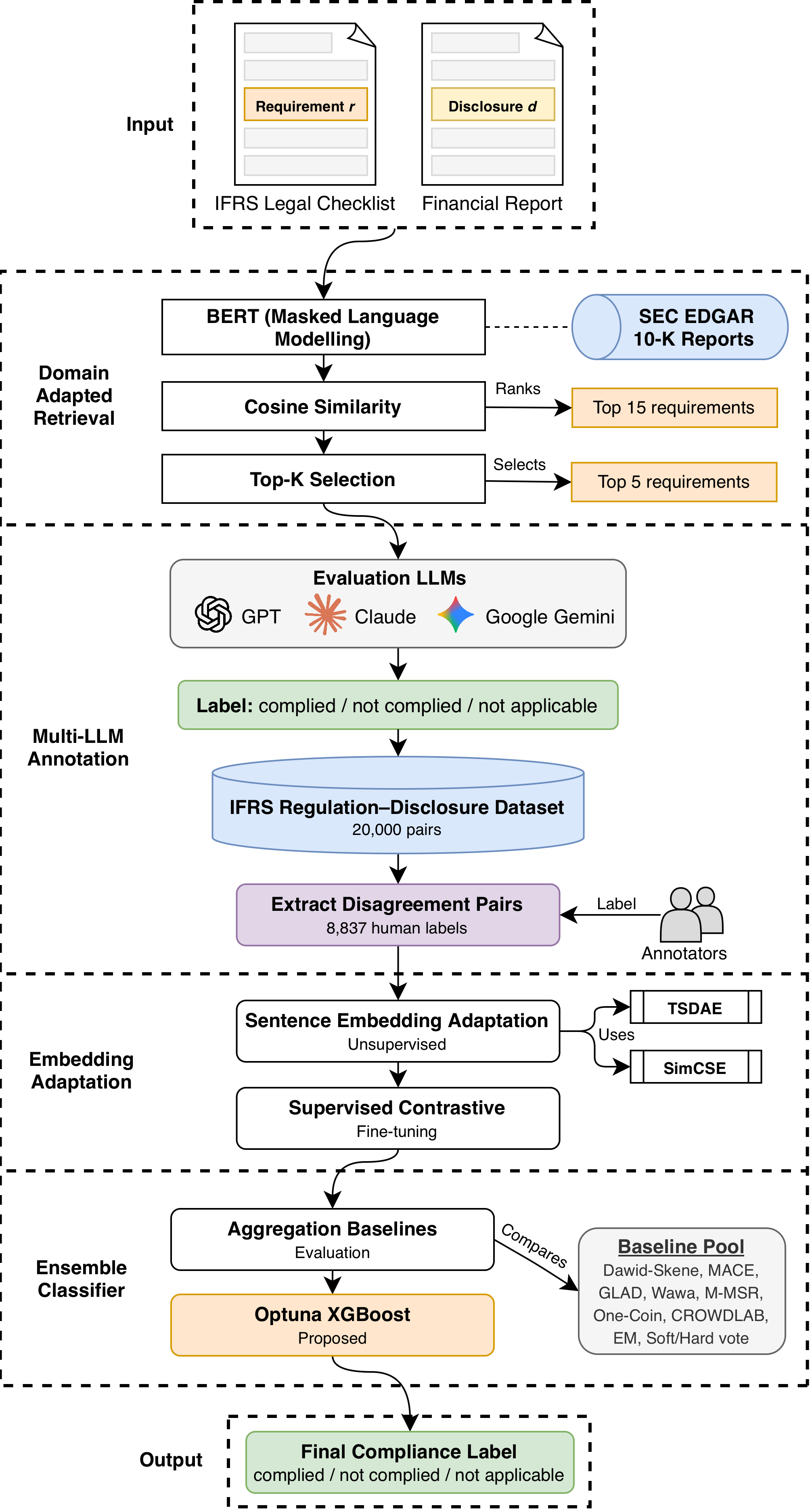}
    \caption{Proposed framework. Top-5 regulations per disclosure are retrieved by an EDGAR-adapted BERT encoder and labeled by a three-LLM panel; 8,837 disagreement items are reconciled with humans. An Optuna-tuned XGBoost meta-learner over LLM probabilities and a frozen SBERT pair embedding predicts the final label, benchmarked against baseline aggregators.}
    \label{fig:framework}
\end{figure}

An end-to-end pipeline for IFRS compliance verification is presented, integrating domain-adapted retrieval, multi-LLM annotation, and a content-conditioned learned aggregator (Figure~\ref{fig:framework}). Candidate regulation-disclosure pairs are constructed by BERT retrieval over an EDGAR-adapted encoder, labeled by a three-LLM ensemble, and reconciled against independent human annotations on the disagreement subset. A three-stage sentence-embedding pipeline yields domain-specific representations, and an Optuna-tuned XGBoost classifier maps the LLM probability panel together with a frozen SBERT embedding of the pair to a final label, benchmarked against a pool of unsupervised aggregation baselines. Each stage is detailed in the subsections that follow.

\subsection{Task Definition and Dataset Construction}
\label{sec:task-definition}

Compliance verification is formalized as a three-way classification problem. Given a pair $(r, d)$ consisting of an IFRS regulation paragraph $r$ and a financial statement disclosure paragraph $d$, the goal is to assign a label $y \in \{\textsc{Complied}, \textsc{Not Complied}, \textsc{Not Applicable}\}$ indicating whether $d$ satisfies the requirements specified in $r$. Concretely, \textsc{Complied} means the requirement applies to the disclosure's topic and $d$ satisfies it; \textsc{Not Complied} means the requirement applies to the disclosure's topic but $d$ alone does not satisfy it, although other passages elsewhere in the filing may satisfy it (the label characterizes the passage-level pairing $(r, d)$, not company-level compliance); and \textsc{Not Applicable} means the requirement does not apply to the disclosure's topic (e.g., a leases requirement evaluated against a revenue disclosure).

The regulation-side corpus consists of 3,489 paragraphs extracted via LlamaParse from the official IASB published PDFs of all 18 active IFRS standards (IFRS 1--3, 5--19). The disclosure-side corpus consists of 39,404 paragraphs from 329 audited annual financial statements filed via SEDAR+\footnote{\url{https://www.sedarplus.ca/}} (fiscal years 2022--2025); jurisdiction, sector composition, issuer-size profile, and selection rationale are detailed in Appendix~\ref{app:corpus}, with the full parsing pipeline in Appendix~\ref{app:llamaparse}.

To construct candidate pairs, all paragraphs are embedded using the domain-adapted BERT encoder pre-trained via MLM on SEC EDGAR 10-K filings, and the top-15 regulations per disclosure are retrieved by cosine similarity. Retrieved scores range from 0.80 to 0.94, confirming high semantic overlap between the two corpora. From the 39,404 disclosures, 4,000 are randomly selected (fixed seed), and the top-5 regulation matches per disclosure are retained by similarity rank, yielding 20,000 pairs over 1,981 distinct regulations from all 18 standards and 4,000 distinct disclosures (statistics in Appendix~\ref{app:dataset-stats}). Two of these pairs remained unresolved after human review (\S\ref{sec:human-annotation}) and are excluded, leaving 19,998 labeled pairs. The top-5 cutoff balances coverage against annotation cost and makes the benchmark conditional on retrieval: it evaluates verification of retrieved candidates, not open retrieve-and-verify. Unretrieved pairs carry no labels. The resulting dataset, labeled by the multi-LLM ensemble described in Section~\ref{sec:llm-annotation}, is imbalanced: \textsc{Not Applicable} accounts for 11,759 pairs (58.8\%), \textsc{Not Complied} for 7,369 (36.8\%), and \textsc{Complied} for only 872 (4.4\%). The imbalance is partly structural (retrieved regulations are topically adjacent but mostly do not govern the disclosure) and partly inherent (one paragraph rarely satisfies a requirement in full). The distribution motivated class-weighted training and stratified evaluation throughout.

\subsection{Multi-LLM Ensemble Annotation}
\label{sec:llm-annotation}

Three frontier LLMs from three independent model families are deployed as parallel annotators: GPT-4.1 \cite{openai_gpt41_2025} from OpenAI, Claude Sonnet 4 \cite{anthropic_claude4_2025} from Anthropic, and Gemini 2.5 Pro \cite{google_gemini25_2025} from Google DeepMind. Each is competitive on financial-reasoning and instruction-following benchmarks individually, but published audits show that no single frontier model is uniformly reliable across compliance categories \cite{berger_towards_2023, fotoh_exploring_2025}, and the per-class analysis (Appendix~\ref{app:per-class}) confirms complementary failure modes: GPT-4.1 over-predicts \textsc{Not Complied}, Claude is balanced but conservative on the minority \textsc{Complied} class, and Gemini is the strongest single model on \textsc{Complied} and \textsc{Not Applicable} yet degrades on \textsc{Not Complied}. All three are queried at temperature $T{=}0$ for deterministic outputs. Each model receives a shared system prompt specifying an expert IFRS auditor role (Appendix~\ref{app:prompt-compliance}) and outputs a structured JSON containing a discrete label and a probability vector $\mathbf{p} = [p_{\text{yes}}, p_{\text{no}}, p_{\text{na}}]$ over the three classes. The prompt queries each class probability independently in $[0, 1]$ rather than enforcing a softmax over the three classes, so each returned $\mathbf{p}$ is L1-normalized to satisfy $\sum_k p_k = 1$ before downstream aggregation, calibration, or feature extraction. Predictions are aggregated via hard vote (majority rule) and soft vote (argmax of averaged probability vectors). Of the 20,000 items, 55.8\% receive unanimous agreement, 44.0\% majority agreement, and only 0.2\% full three-way disagreement.

\subsection{Human Annotation for Ground Truth}
\label{sec:human-annotation}

Independent human annotations on the non-unanimous subset are collected through a purpose-built platform (Appendix~\ref{app:platform}). Three junior IFRS analysts label each pair blind to LLM predictions and to one another, supervised by two senior IFRS specialists who designed the protocol and adjudicated escalated cases. Each analyst handles a disjoint partition of roughly 3,000 partial-agreement items; a fixed 163-item overlap slice, which includes all 44 items on which the three LLMs disagree three ways, is multiply annotated (161 by all three analysts, 2 by two). On the 161, mean pairwise agreement is 65.8\% (Fleiss' $\kappa = 0.39$), modest, but measured on items the three LLMs already disagree on. The analysts are unanimous on 81 of these items, split two to one on 75, and split three ways on 5, where no majority exists. Two-to-one splits take the majority label. All 80 split items and the 21 items flagged as difficult were referred to the two senior specialists for joint review, which agreed a final label for three of the five three-way items; the other two remained unresolved when labeling closed and are excluded from the released labels. In total, 8,837 items are annotated (44.2\% of the dataset): 73.9\% Not Applicable, 15.8\% Not Complied, 10.3\% Complied. The minority \textsc{Complied} share is 2.3$\times$ higher in this disagreement subset than in the full LLM-labeled set (4.4\%, \S\ref{sec:task-definition}), confirming that LLM disagreement concentrates around the under-represented positive class (partition details in Appendix~\ref{app:partition}).

\subsection{Domain-Adapted Sentence Embedding Pipeline}
\label{sec:embedding-pipeline}

\paragraph{Stage 1: MLM Pre-training.}
The encoder is initialized from \texttt{bert-base-cased} and further pre-trained with MLM on the Management Discussion and Analysis sections of SEC EDGAR 10-K filings (1993–2020), chunked into overlapping 512-token segments with 64-token stride and 15\% masking. Training uses 3 epochs, batch size 16, learning rate $5\times10^{-5}$, weight decay 0.01, and fp16 mixed precision. The cased tokenizer is retained to preserve capitalization-sensitive acronyms (IFRS, IAS, GAAP).

\paragraph{Stage 2A: TSDAE.}
The encoder compresses a 60\% token deleted input into a 768-dimensional vector via mean pooling; a decoder with tied weights reconstructs the original sentence, forcing holistic semantic encoding. Training uses 42,893 unlabeled sentences for 1 epoch with batch size 8 and learning rate $3\times10^{-5}$.

\paragraph{Stage 2B: SimCSE.}
Positive pairs are formed by passing each sentence through the encoder twice with independently sampled dropout masks; all other batch members serve as negatives under InfoNCE loss. Training uses the same corpus for 1 epoch with batch size 128 (127 in-batch negatives per anchor), learning rate $3 \times 10^{-5}$, and temperature $\tau{=}0.05$.

\paragraph{Stage 3: Supervised Contrastive Fine-Tuning.}
Both Stage 2 checkpoints are independently fine-tuned on 11,161 unanimous-agreement items. Pairs labeled \textsc{Complied} (547 pairs) or \textsc{Not Complied} (4,383 pairs) form 4,930 positive training pairs. After a 90/10 split, training uses MultipleNegativesRankingLoss, batch size 16, learning rate $2 \times 10^{-5}$, $\tau{=}0.05$, and 4 epochs.

\subsection{Learned Ensemble Classifier}
\label{sec:learned-en-classifier}

This work frames aggregation as \emph{stacked generalization} \cite{wolpert_stacked_1992, sagi_ensemble_2018}: the three LLMs of \S\ref{sec:llm-annotation} are level-0 base learners emitting per-class probabilities, and an XGBoost \cite{chen_xgboost_2016} classifier is the level-1 meta-learner mapping those probabilities, together with a frozen content representation, to the final label; unlike hard/soft vote or unsupervised competence models (Dawid--Skene, MACE, CROWDLAB), it is trained on human ground truth and also reads the input's domain-adapted embedding, so the decision depends on \emph{what the pair says} as well as on what the panel said. The 1,565-dim input concatenates a 29-dim panel summary (nine raw class probabilities, per-model Shannon entropies, hard-vote count, soft-vote confidence, max per-pair confidence, five binary agreement-pattern indicators, per-class max--min spread and standard deviation across models, three pairwise symmetric KL divergences) and the 1,536-dim frozen SBERT embedding \cite{reimers_sentence_bert_2019} of the regulation--disclosure pair from the TSDAE Stage-2A (unsupervised, domain-adapted) checkpoint (\S\ref{sec:embedding-pipeline}); only the meta-learner is trained. Gradient-boosted trees fit this setting on four counts: bounded panel statistics and the dense embedding are handled in native scales without standardization; at 8,837 items under 5-fold CV, regularized boosting dominates end-to-end deep alternatives (Appendix~\ref{app:router-ablations}, vs.\ fine-tuned BERT and confidence-weighted neural classifiers); per-sample weights, exposed as three free class weights, push capacity onto the minority \textsc{Complied}/\textsc{Not Complied} classes rather than collapsing to NA; and L1/L2 leaf penalties with depth and shrinkage fit the 1,565-dim input on $\sim$7,000 items per fold without overfit. Optuna \cite{akiba_optuna_2019} with the Tree-structured Parzen Estimator
(TPE) sampler then selects hyperparameters over 100 trials, each scored by 5-fold stratified CV weighted F1, jointly tuning tree-structure parameters (estimators, depth, learning rate, subsample, column subsample, min child weight, $\gamma$, $\alpha$, $\lambda$) and three class weights; the chosen configuration is 777 trees of depth 7, learning rate 0.038, weights $(w_{\textsc{NA}}, w_{\textsc{NO}}, w_{\textsc{YES}}){=}(1.0, 3.5, 7.1)$, without which the meta-learner collapses to an NA-dominant policy with W-F1 below 0.75. Which LLM to trust is content-dependent, not confidence-dependent.

\paragraph{Splits and leakage.} All classifier predictions are out-of-fold under stratified 5-fold cross-validation on the 8,837 human-labeled items, which share no pair with the 11,161 unanimous items used for contrastive training; the classifier's Stage-2A encoder sees no labels. Folds are stratified by label, not grouped by disclosure, so the evaluation measures generalization to unseen pairs, not unseen filings.

\subsection{Evaluation Protocol}
\label{sec:eval-protocol}

Four aspects of the framework are evaluated: retrieval quality, calibration diagnostics, aggregation comparison, and IAA.

\paragraph{Retrieval Evaluation.}
Recall@$K$ for $K \in \{1, 5, 10, 15\}$ is reported separately on seen regulations (present in Stage 3 training) and unseen regulations (held out entirely), isolating in-domain retrieval from zero-shot generalization.

\paragraph{Calibration Diagnostics.}
Per-model accuracy and Cohen's $\kappa$ are computed against the 8,837 human labels for each LLM, hard vote, soft vote, and EM-calibrated predictions (algorithm in Appendix~\ref{app:em-algorithm}). Calibration uses ECE (15 bins).

\paragraph{Aggregation Benchmark.}
Eleven aggregation strategies are evaluated on the 8,837-item set and compared on accuracy, weighted F1, macro F1, and Cohen's $\kappa$ against the learned ensemble classifier described in Section \ref{sec:learned-en-classifier}.

\section{Experiments}
\label{sec:experiments}

\subsection{Retrieval Results}
\label{sec:retrieval-results}

The five model checkpoints from the embedding pipeline (\S\ref{sec:embedding-pipeline}) are evaluated on the retrieval task: given a regulation, all 39,404 disclosure candidates are ranked by cosine similarity and human-labeled relevant disclosures are checked for appearance in the top $K$ positions. All models are evaluated on independent human annotations (8,837 items) collected separately from the LLM labels used for supervised training. Recall@$K$ and Mean Reciprocal Rank (MRR) are reported with 95\% bootstrap confidence intervals from 1,000 resamples, separately for seen regulations (present in supervised training with different disclosure pairings) and unseen regulations (held out entirely for zero-shot evaluation). Table \ref{tab:retrieval} reports the headline results; full results including MAP@15 and NDCG@15 for all five checkpoints are in Appendix~\ref{app:retrieval}; per-stage Hyperparameters are in Appendix~\ref{app:hyperparams}.

\begin{table}[t]
\centering\small
\setlength{\tabcolsep}{4pt}
\begin{tabular}{lccc}
\toprule
\textbf{Model} & R@5 & R@15 & MRR \\
\midrule
\multicolumn{4}{l}{\emph{Seen Regulations} ($n{=}530$)} \\
MLM only             & 8.9  & 20.9 & 0.060 \\
\quad + TSDAE        & 7.7  & 19.8 & 0.053 \\
\quad + SimCSE       & 7.4  & 15.8 & 0.046 \\
\quad + TSDAE + Sup  & \textbf{11.9} & \textbf{27.5} & \textbf{0.080} \\
\quad + SimCSE + Sup & 11.3 & 26.4 & 0.073 \\
\midrule
\multicolumn{4}{l}{\emph{Unseen Regulations} ($n{=}106$)} \\
MLM only             & 11.3 & 28.3 & 0.086 \\
\quad + TSDAE        & 12.3 & 26.4 & 0.069 \\
\quad + SimCSE       & 12.3 & 27.4 & 0.078 \\
\quad + TSDAE + Sup  & 18.9 & 41.5 & 0.122 \\
\quad + SimCSE + Sup & \textbf{20.8} & \textbf{45.3} & \textbf{0.137} \\
\bottomrule
\end{tabular}
\caption{Retrieval performance on (3,489 regulations, 39,404 disclosures).}
\label{tab:retrieval}
\end{table}

\paragraph{Supervised training is the decisive step.}
Both unsupervised methods alone provide marginal improvement over the MLM baseline, with Recall@5 below 13\% for both splits, consistent with Biesner et al.~\cite{biesner_zero-shot_2022}. After Stage 3 fine-tuning, all metrics improve substantially. SimCSE + Supervised reaches 20.8\% unseen R@5 and 45.3\% unseen R@15; TSDAE + Supervised leads on seen regulations (R@5 = 11.9\%, MRR = 0.080). The two models are complementary: TSDAE's denoising objective produces representations better suited to familiar regulation patterns, while SimCSE's contrastive objective generalizes more effectively to novel regulations.

% Leaderboard Figure
\begin{figure}[t]
\centering
\includegraphics[width=1\linewidth]{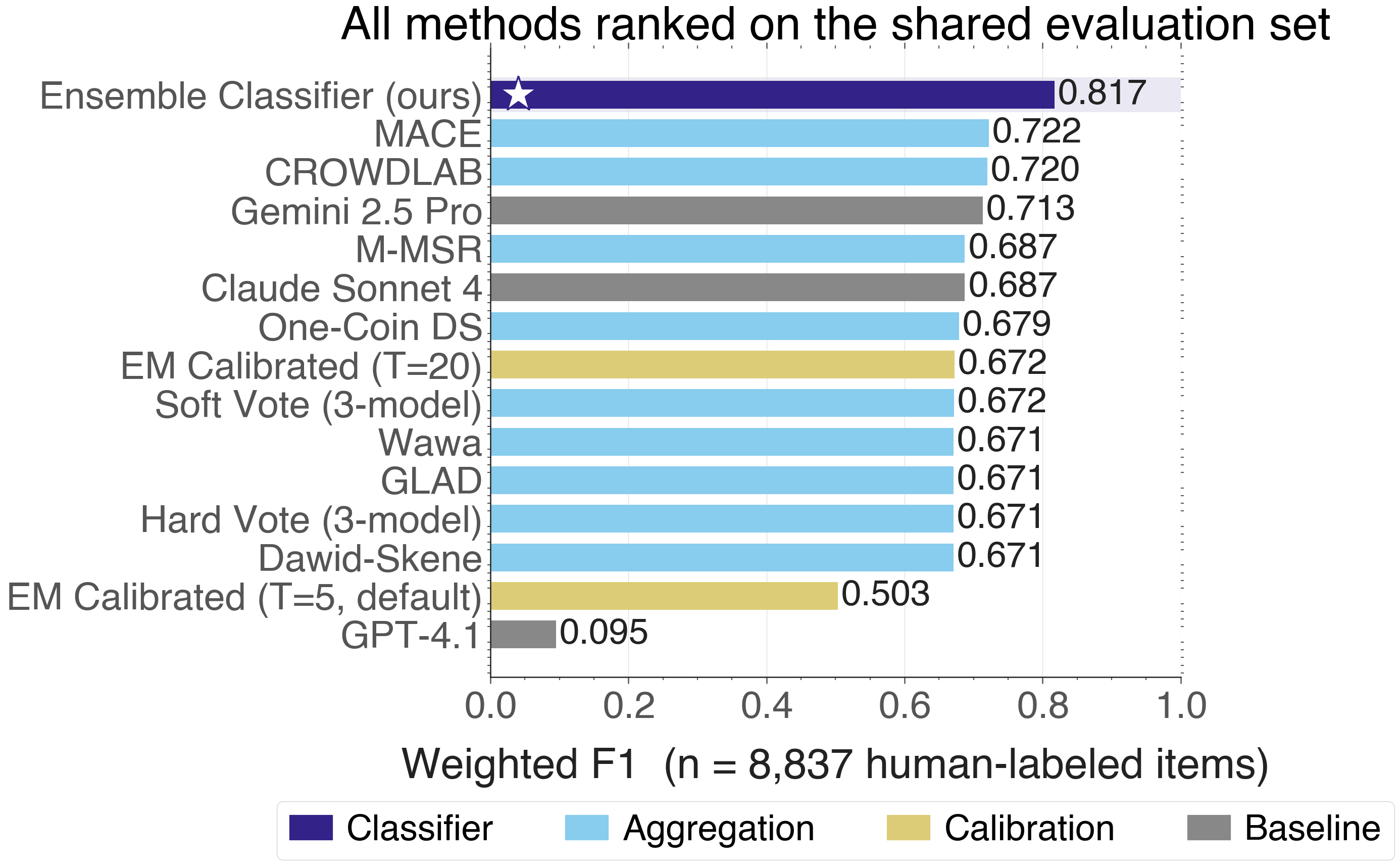}
\caption{Method leaderboard: all 15 evaluated methods (three LLMs, eleven aggregators, the proposed classifier) ranked by weighted F1 against 8,837 independent human annotations. The learned ensemble classifier (starred) reaches 0.817, exceeding the best single LLM (Gemini, 0.713) and the best unsupervised aggregator (MACE, 0.722).}
\label{fig:leaderboard}
\end{figure}

\subsection{Classification Benchmark and Learned Ensemble Classifier}
\label{sec:aggregation-results}

Table~\ref{tab:classification} reports results across individual LLMs, representative aggregation methods, and the learned ensemble classifier; per-class metrics and the full 11-strategy comparison are in Appendix~\ref{app:per-class} and Appendix~\ref{app:aggregation-extended} respectively.

\begin{table}[!t]
\centering\footnotesize
\setlength{\tabcolsep}{4pt}
\begin{tabular}{>{\raggedright\arraybackslash}p{3.5cm}cccc}
\toprule
\textbf{Method} & \textbf{Acc} & \textbf{W-F1} & \textbf{$\kappa$} & \textbf{ECE} \\
\midrule
\multicolumn{5}{l}{\emph{\textbf{Individual LLMs}}} \\
GPT-4.1 \cite{openai_gpt41_2025}                & 0.183 & 0.095 & 0.019 & 0.811 \\
Claude Sonnet~4 \cite{anthropic_claude4_2025}   & 0.672 & 0.687 & 0.256 & 0.211 \\
Gemini 2.5 Pro \cite{google_gemini25_2025} & 0.737 & 0.713 & 0.278 & 0.215 \\
\midrule
\multicolumn{5}{l}{\emph{\textbf{Aggregation}}} \\
Hard Vote                                        & 0.641 & 0.671 & 0.254 & --    \\
Soft Vote                                        & 0.642 & 0.672 & 0.255 & 0.290 \\
Dawid-Skene \cite{dawid_maximum_1979}            & 0.641 & 0.671 & 0.254 & --    \\
One-Coin DS \cite{zhang_spectral_em_2016}        & 0.668 & 0.679 & 0.238 & --    \\
EM Calibrated (T=5, default)                     & 0.456 & 0.503 & 0.149 & 0.461 \\
EM Calibrated (T=20) \cite{guo_calibration_2017} & 0.647 & 0.672 & 0.245 & 0.292 \\
Wawa \cite{ustalov_crowdkit_2024}                & 0.642 & 0.671 & 0.254 & --    \\
M-MSR \cite{ma_mmsr_2020}                        & 0.672 & 0.687 & 0.256 & --    \\
GLAD \cite{whitehill_whose_2009}                 & 0.642 & 0.671 & 0.254 & --    \\
CROWDLAB \cite{goh_crowdlab_2022}                & 0.791 & 0.720 & 0.340 & --    \\
MACE \cite{hovy_learning_2013}                   & 0.748 & \textbf{0.722} & 0.312 & --    \\
\midrule
\multicolumn{5}{l}{\emph{\textbf{Proposed}}} \\
\textbf{Optuna XGBoost (proposed)}                   & \textbf{0.824} & \textbf{0.817} & \textbf{0.556} & \textbf{0.068} \\
\bottomrule
\end{tabular}
\caption{Classification benchmark on 8,837 disagreement items.}
\label{tab:classification}
\end{table}

% Confusion Matrix Figure
\begin{figure}[t]
\centering
\includegraphics[width=1\linewidth]{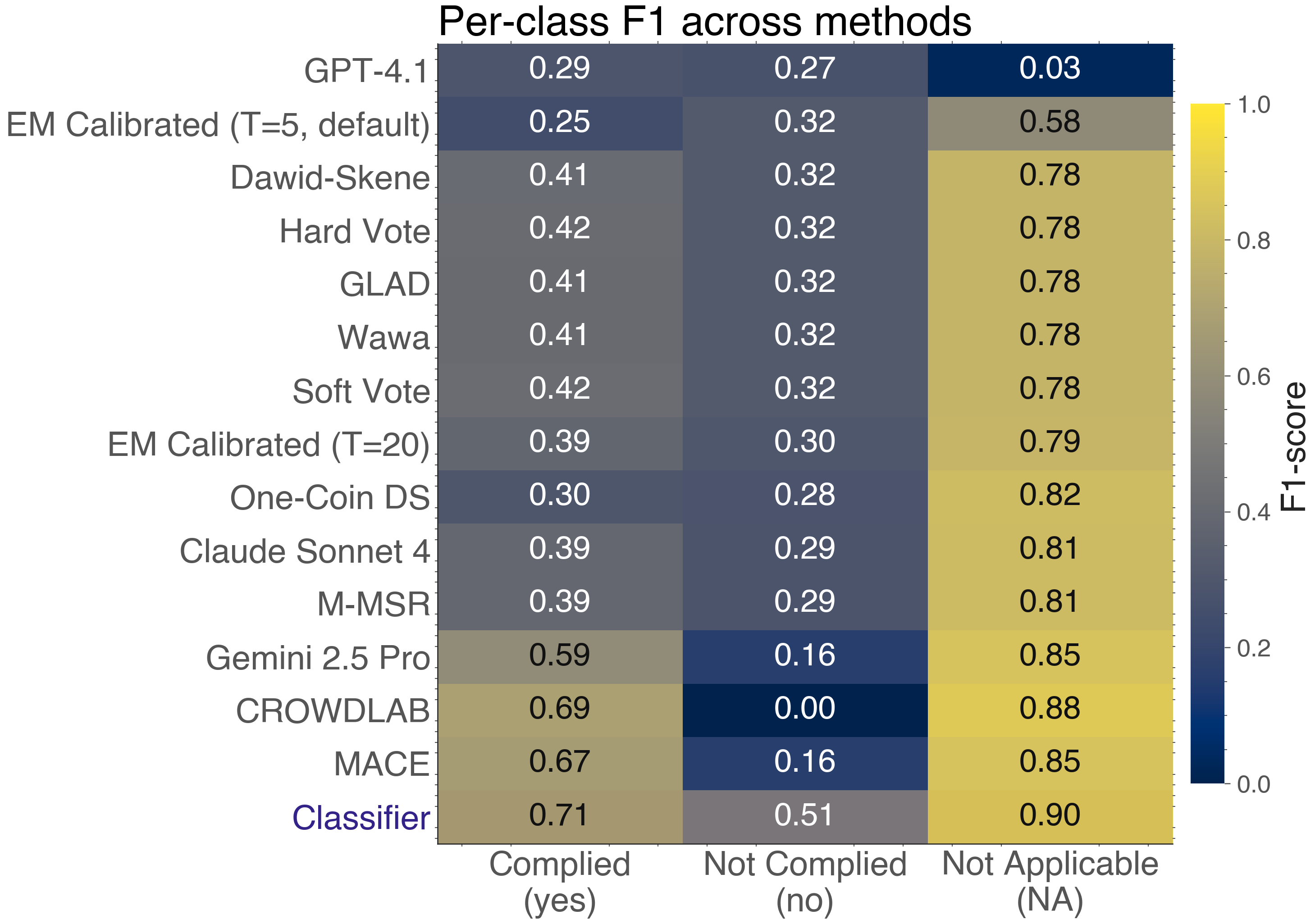}
\caption{Per-class F1 across all 15 evaluated methods on 8,837 human-annotated items.}
\label{fig:perclass}
\end{figure}

\paragraph{Aggregation partially recovers signal.} 
Figure \ref{fig:leaderboard} ranks all methods by weighted F1. GPT-4.1 exhibits severe class-level bias on the human-annotated disagreement subset: 95.5\% of its predictions there are \textsc{Not Complied}, yielding near-chance agreement with humans ($\kappa=0.019$, Table \ref{tab:classification}). This biased contribution pulls hard vote (0.671 W-F1) and soft vote (0.672 W-F1) below the best single model, Gemini, at 0.713 W-F1. MACE \cite{hovy_learning_2013} is the strongest aggregator at 0.722 W-F1. Its single scalar reliability parameter is more stable at this panel size than Dawid-Skene's full confusion matrix: with only three annotators the 9-parameter per-model matrix is undetermined, and the M-step collapses to majority voting, explaining why Dawid-Skene ties hard vote exactly at 0.671 (Table \ref{tab:classification}). Several other aggregators (GLAD, Wawa, One-Coin DS, M-MSR) fall in the same 0.671–0.687 band: extra parameters do not help with three annotators.

\paragraph{CROWDLAB narrows but does not close the gap to the proposed method.}
CROWDLAB \cite{goh_crowdlab_2022} achieves the best accuracy (0.791) and Cohen's kappa (0.340) among unsupervised aggregators (Table \ref{tab:classification}), with W-F1 (0.720) close to MACE. The split between W-F1 and accuracy indicates a deeper failure mode: CROWDLAB attains 0.88 F1 on \textsc{NOT APPLICABLE} and 0.69 F1 on \textsc{COMPLIED}, but 0.00 F1 on \textsc{NOT COMPLIED} (Figure \ref{fig:perclass}). CROWDLAB's classifier-augmented consensus drives the model to abandon the minority class for \textsc{NOT APPLICABLE}.

\paragraph{EM calibration's value is diagnostic.}
Under the adopted bound $T_{\max}{=}20.0$, learned temperatures correctly rank annotator reliability (Anthropic $T{=}0.55$, Gemini $T{=}5.73$, OpenAI $T{=}14.36$), with no model pinned at the bound; the calibrated ensemble reaches 0.672 W-F1 and ECE 0.292, matching soft vote. Per-model ECE improves substantially for the two over-confident annotators (OpenAI $0.811{\to}0.508$; Gemini $0.215{\to}0.123$) but rises slightly for the already near-calibrated Anthropic ($0.211{\to}0.276$), since EM over-sharpens its raw outputs in service of the consensus fit (Appendix~\ref{app:reliability}). Under the textbook bound $T_{\max}{=}5.0$, OpenAI and Gemini both pin at 5.0 and ensemble W-F1 collapses to 0.503 (Table~\ref{tab:classification}; bound sensitivity in Appendix~\ref{app:em-audit}). Either way, EM does not exceed unsupervised aggregation: its practical value is identifying which models to trust, not directly producing labels.

\paragraph{The learned ensemble classifier closes 41.9\% of the oracle gap.}
A per-item oracle selecting the correct model per pair reaches 0.949 W-F1, a 22.7-point gap above MACE that motivates a content-conditioned approach. The Optuna-tuned XGBoost classifier (\S\ref{sec:learned-en-classifier}) reaches 0.817 W-F1, beating every aggregation baseline by at least 9.5 points, with gains concentrated on the minority \textsc{Not Complied} and \textsc{Complied} classes (Figure~\ref{fig:perclass}). Probability features alone saturate at 0.727 W-F1; adding the frozen 1,536-dimensional SBERT embedding \cite{reimers_sentence_bert_2019} of the regulation--disclosure pair contributes $+$8.1 points and hyperparameter search a further 0.9 (Appendix~\ref{app:router-ablations}), showing that which model to trust is content-dependent.

\section{Conclusion}
\label{sec:conclusion}
This paper presents a multi-stage framework for IFRS compliance verification and reports three findings. First, the domain-adapted embedding pipeline (BERT$\to$TSDAE/SimCSE$\to$supervised contrastive) reaches 45.3\% Recall@15 on unseen regulations against 39,404 candidates, confirming industrial-scale transfer. Second, training the contrastive stage on 4,930 LLM-unanimous pairs yields overlapping 95\% bootstrap confidence intervals with training on 1,616 human-annotated pairs, supporting LLM-unanimous labels as a practical weak-supervision alternative in this setting. Third, naive aggregation is fragile under single-model bias: GPT-4.1's 95.5\% ``Not Complied'' rate on the disagreement subset pulls hard and soft voting (0.671/0.672 W-F1) below the best single model (Gemini, 0.713), whereas an Optuna-tuned XGBoost over frozen embeddings and LLM probability features reaches 81.7\% W-F1, closing 41.9\% of the per-item oracle gap.

The dataset, checkpoints, classifier code, and annotation platform will be released to support further work on multi-LLM annotation, financial NLP, and reliability-aware ensembles.

\section*{Limitations}
No claim of universality is made. The framework is evaluated on a single domain (IFRS-aligned annual financial statements), a single language (English), a single class taxonomy (\textsc{Complied} / \textsc{Not Complied} / \textsc{Not Applicable}), and a fixed three-LLM annotator panel. Within that scope, the following caveats qualify the reported findings.

\paragraph{Paragraph-level scope.}
Labels are assigned to paragraph-level pairings $(r, d)$ rather than to whole filings. A pair marked \textsc{Not Complied} only indicates that disclosure paragraph $d$ alone does not satisfy regulation $r$; another passage in the same filing may, and a single \textsc{Complied} pair establishes only that the requirement is met somewhere, not that no contradiction exists elsewhere. Deriving a filing-level verdict therefore requires an additional aggregation step over disclosure paragraphs, which is outside the reported evaluation.

\paragraph{Retrieval candidate pre-selection.}
Candidate regulation--disclosure pairs are pre-selected by a domain-adapted BERT (MLM-fine-tuned on EDGAR) using top-5 cosine ranking; pairs scored below rank 5 by this model are absent from the evaluation set. The relevant pairings this excludes cannot be counted from the benchmark, since unretrieved pairs carry no labels; the pre-selection encoder's own Recall@15 of 20.9\% (seen) and 28.3\% (unseen regulations; Table~\ref{tab:retrieval}, MLM only, ranking disclosures for a regulation) suggests the exclusion is substantial. A stratified random sample from the full $3,489 \times 39,404$ pair space would yield an unbiased benchmark; this benchmark is conditional on that model's relevance prior.

\paragraph{Label reliability.}
Human labels cover only the disagreement subset; the 11,161 unanimous items carry LLM labels that no human has validated, so any score on them measures agreement with the LLM label, not accuracy (Appendix~\ref{app:router-ablations}), and the 0.817 on the disagreement subset should not be read as a bound on performance over the full benchmark. Inter-annotator agreement on the overlap slice is modest ($\kappa = 0.39$), and most human items are singly annotated, so individual labels on the hardest pairs carry noise that majority voting and senior review reduce but do not remove.

\paragraph{Commercial audit tools.}
No comparison with commercial audit products is reported: the AI tools of large audit firms are proprietary, expose no public interface, and report no paragraph-level IFRS metrics, so a reproducible comparison is not currently possible. Once released, the benchmark is intended to make such comparisons possible for their developers.

\paragraph{Small annotator panel.}
Three LLMs is a small ensemble. Extensions to 5--10 annotators would exercise Dawid--Skene and related matrix-estimation methods more fairly and may change the ranking of aggregation methods, and consequently the gain attributable to the learned ensemble classifier.

\paragraph{Class imbalance.}
The dataset is heavily skewed (4.4\% \textsc{Complied} / 36.8\% \textsc{Not Complied} / 58.8\% \textsc{Not Applicable}). The learned ensemble classifier was trained and evaluated on this distribution; performance on regulatory regimes with materially different applicability rates is not verified.

\paragraph{Corpus and transfer.}
The disclosures come from an unstratified convenience sample of 329 SEDAR+ filings, weighted toward small- and mid-cap Canadian issuers and the mining sector (Appendix~\ref{app:corpus}); results may not transfer to large-cap or other-sector reporting without re-evaluation. Only two components are IFRS-specific, the domain-adapted encoder and the annotation prompt; the stacking architecture is not. Transfer to other regimes (US GAAP, Basel III), which use different document conventions and taxonomies, to non-English filings, and to finer label taxonomies is untested.

\paragraph{Commercial LLM dependency and reproducibility cost.}
The annotation pipeline relies on three commercial APIs (GPT-4.1, Claude Sonnet~4, Gemini 2.5 Pro). Specific model versions may be deprecated; classifier features derived from their probability outputs may not transfer to successor models without re-tuning. Open-weight annotators are not evaluated here. The classifier accepts any annotator from which per-class scores can be extracted but must be refit when the panel changes; the frozen content embedding, which carries most of the gain over probability features (Appendix~\ref{app:router-ablations}), does not depend on the panel. Deployment cost is dominated by this panel: the classifier adds one boosted-tree pass per pair, while the three LLM calls, independent and parallelizable, cost at least USD 0.003 per pair at list prices, about USD 60 for the 20,000-pair corpus (Appendix~\ref{app:cost}).

% \section*{Acknowledgments}

\bibliography{custom}

\appendix

\section*{Appendix}
\addcontentsline{toc}{section}{Appendix}

\section{Disclosure Corpus Provenance}
\label{app:corpus}

The 329 audited annual financial statement (AFS) PDFs that form the disclosure side of the benchmark (Section~\ref{sec:task-definition}) are predominantly Canadian, obtained from SEDAR+ (the public regulatory portal operated by the Canadian Securities Administrators, on which reporting issuers listed on Canadian exchanges including TSX, TSX-V, CSE, and NEO are required to file their continuous-disclosure documents), with a small number of cross-listed U.S. 10-K reports (12 filings) and Quebec-origin French-language filings (3 filings); English-language Canadian filings constitute the bulk of the set. Fiscal year-ends span 2022 through 2025, with 93\% of filings reporting on fiscal year 2025; the three most common quarter-end dates are 30 June (159 filings), 31 July (80 filings), and 31 August (28 filings), reflecting the typical Canadian junior-issuer fiscal calendar. The corpus is heavily weighted toward small- and mid-cap reporters, including a substantial population of capital pool companies and junior exploration issuers; large-cap issuers such as chartered banks and senior producers are present but proportionally rare. Mining and resource exploration is the single largest sector (approximately 30\% of the corpus), with secondary representation from capital pool companies, financial-holding entities, and biotechnology and life sciences, and a long tail of consumer, industrial, and real-estate issuers. The 329 filings were assembled as an un-stratified convenience sample of publicly available SEDAR+ filings within the data collection window, retained without sector, size, or auditor-firm stratification so that the benchmark preserves the natural distribution of disclosure language a compliance verifier encounters in practice rather than enforcing artificial balance.

\section{Dataset Statistics}
\label{app:dataset-stats}

The 19,998 labeled pairs combine 1,981 distinct IFRS regulations, drawn from all 18 standards, with 4,000 distinct disclosures; each disclosure contributes its top-5 retrieved regulations. Table~\ref{tab:per-standard} breaks the pairs down by standard and Table~\ref{tab:subset-classes} by label source. The distribution in \S\ref{sec:task-definition} is the three-LLM hard vote over all pairs; the tables here instead use the human label wherever one exists, which is why \textsc{Complied} is larger here (1,461 against 872): LLM disagreement concentrates on the positive class. Coverage is uneven across standards, from 2,556 pairs for IFRS~2 to 334 for IFRS~19, and IFRS~14 (regulatory deferral accounts) has no \textsc{Complied} pair at all.

\begin{table}[h]
\centering\footnotesize
\setlength{\tabcolsep}{3pt}
\begin{tabular}{lrrrrrr}
\toprule
\textbf{Standard} & \textbf{Regs} & \textbf{Pairs} & \textbf{C} & \textbf{NC} & \textbf{NA} & \textbf{Human \%} \\
\midrule
IFRS 1 & 97 & 893 & 32 & 185 & 676 & 61 \\
IFRS 2 & 138 & 2,556 & 297 & 798 & 1,461 & 45 \\
IFRS 3 & 140 & 1,192 & 67 & 207 & 918 & 46 \\
IFRS 5 & 61 & 607 & 26 & 101 & 480 & 45 \\
IFRS 6 & 25 & 672 & 113 & 281 & 278 & 35 \\
IFRS 7 & 227 & 2,395 & 26 & 985 & 1,384 & 42 \\
IFRS 8 & 43 & 335 & 13 & 51 & 271 & 51 \\
IFRS 9 & 127 & 2,167 & 306 & 696 & 1,165 & 36 \\
IFRS 10 & 111 & 933 & 71 & 185 & 677 & 38 \\
IFRS 11 & 47 & 447 & 31 & 76 & 340 & 51 \\
IFRS 12 & 84 & 489 & 22 & 94 & 373 & 51 \\
IFRS 13 & 88 & 994 & 74 & 490 & 430 & 39 \\
IFRS 14 & 56 & 848 & 0 & 143 & 705 & 68 \\
IFRS 15 & 176 & 1,129 & 63 & 297 & 769 & 40 \\
IFRS 16 & 158 & 1,391 & 113 & 368 & 910 & 42 \\
IFRS 17 & 222 & 1,390 & 73 & 358 & 959 & 36 \\
IFRS 18 & 140 & 1,226 & 94 & 373 & 759 & 48 \\
IFRS 19 & 41 & 334 & 40 & 89 & 205 & 68 \\
\midrule
Total & 1,981 & 19,998 & 1,461 & 5,777 & 12,760 & 44 \\
\bottomrule
\end{tabular}
\caption{Pairs per IFRS standard. Labels are human where available, otherwise the unanimous LLM label (C = \textsc{Complied}, NC = \textsc{Not Complied}, NA = \textsc{Not Applicable}). Human \% is the share of the standard's pairs in the human-annotated disagreement subset.}
\label{tab:per-standard}
\end{table}

\begin{table}[h]
\centering\footnotesize
\begin{tabular}{lrr}
\toprule
 & \textbf{Unanimous} & \textbf{Human} \\
\midrule
Pairs & 11,161 & 8,837 \\
\textsc{Complied} & 547 & 914 \\
\textsc{Not Complied} & 4,383 & 1,394 \\
\textsc{Not Applicable} & 6,231 & 6,529 \\
\bottomrule
\end{tabular}
\caption{Class distribution by label source. The two subsets share no pair; the 2 remaining pairs are three-way analyst splits left unresolved and excluded from the human labels. Regulation texts average 44.3 words (55.1 BERT wordpieces) and disclosures 27.6 words (34.0 wordpieces).}
\label{tab:subset-classes}
\end{table}

\section{Annotation Cost}
\label{app:cost}

Token usage and per-call latency were not logged during annotation, so Table~\ref{tab:cost} estimates cost from the exact annotation prompt of Appendix~\ref{app:prompt-compliance} at each provider's list price (per million input/output tokens: GPT-4.1 USD 2/8, Claude Sonnet~4 USD 3/15, Gemini 2.5 Pro USD 1.25/10).

\begin{table}[h]
\centering\footnotesize
\begin{tabular}{lr}
\toprule
\textbf{Model} & \textbf{USD per 1,000 pairs} \\
\midrule
GPT-4.1 & 0.89 \\
Claude Sonnet 4 & 1.41 \\
Gemini 2.5 Pro & 0.68 \\
\midrule
Panel total & 2.97 \\
\bottomrule
\end{tabular}
\caption{Estimated annotation cost at list prices: 344 input tokens per call on average (254 of them the system prompt) and 25 output tokens, counted with the o200k tokenizer on the exact prompt over all 20,000 pairs. Excludes structured-output schema overhead and Gemini 2.5 Pro reasoning tokens, so it is a lower bound.}
\label{tab:cost}
\end{table}

\section{Document Parsing Pipeline}
\label{app:llamaparse}

Raw PDF documents, specifically IFRS standards issued by the IASB and corporate annual financial statements filed on SEDAR+, are converted to structured data through a three-stage pipeline: cloud-based PDF parsing, JSON refinement, and LLM-driven checklist extraction. Algorithm~\ref{alg:pipeline} summarizes the full document-to-pair construction process. Initial pairing uses the sentence encoder $\phi$ of Section \ref{sec:embedding-pipeline}.

\subsection{PDF Ingestion via LlamaParse}

Parsing uses LlamaParse \cite{llamaparse2024}, a cloud-based document parser, in its agent-assisted mode. An auxiliary LLM (GPT-4.1-mini) interprets complex layouts (multi-page tables, nested lists, footnote structures) that are pervasive in both IFRS standards and note disclosures. Table~\ref{tab:llamaparse-config} lists the configuration.

\begin{table}[H]
\centering
\small
\begin{tabular}{@{}ll@{}}
\toprule
\textbf{Parameter} & \textbf{Value} \\
\midrule
\texttt{parse\_mode} & \texttt{parse\_page\_with\_agent} \\
\texttt{model} & \texttt{openai-gpt-4-1-mini} \\
\texttt{high\_res\_ocr} & \texttt{True} \\
\texttt{adaptive\_long\_table} & \texttt{True} \\
\texttt{outlined\_table\_extraction} & \texttt{True} \\
\texttt{output\_tables\_as\_HTML} & \texttt{True} \\
\bottomrule
\end{tabular}
\caption{LlamaParse configuration for PDF ingestion.}
\label{tab:llamaparse-config}
\end{table}

\begin{algorithm}[t]
\caption{Document-to-Pair Construction Pipeline}
\label{alg:pipeline}
\begin{algorithmic}[1]
\Require IFRS PDFs $\mathcal{P}_r$, financial statement PDFs $\mathcal{P}_d$
\Ensure Candidate pairs $\mathcal{D} = \{(r_i, d_j)\}$
\For{each $p \in \mathcal{P}_r \cup \mathcal{P}_d$}
    \State $\mathrm{struct}_p \gets \textsc{Refine}(\textsc{LlamaParse}(p))$
\EndFor
\State $\mathcal{R} \gets \bigcup_{p \in \mathcal{P}_r} \textsc{Extract}(\mathrm{struct}_p, \pi_{\mathrm{reg}})$
\State $\mathcal{F} \gets \bigcup_{p \in \mathcal{P}_d} \textsc{Extract}(\mathrm{struct}_p, \pi_{\mathrm{fs}})$
\State $\mathbf{E}_{\mathcal{R}} \gets \phi(\mathcal{R})$
\For{each $d_j \in \mathcal{F}$}
    \State $\mathrm{cand}_j \gets \textsc{TopK}\big(\cos(\phi(d_j), \mathbf{E}_{\mathcal{R}}),\, 15\big)$
    \State $\mathrm{pairs}_j \gets \textsc{Sample}(\mathrm{cand}_j,\, 5)$
\EndFor
\State \Return $\mathcal{D} = \bigcup_j \mathrm{pairs}_j$
\end{algorithmic}
\end{algorithm}

\subsection{JSON Refinement}

LlamaParse returns a verbose JSON structure with rendering metadata. A deterministic refinement step normalises this to a simplified schema: each page becomes an ordered list of typed items (\texttt{heading}, \texttt{paragraph}, \texttt{table}), discarding layout coordinates while preserving document hierarchy and table cell structure.

\subsection{Checklist Extraction and Pair Construction}

Two LLM-driven extraction prompts (to be released with the code) are applied to the refined JSON:
\begin{enumerate}
\item The regulatory checklist extractor processes each IFRS standard, yielding 3,489 regulation items with paragraph-level references and verbatim requirement text.
\item The financial statement extractor processes each financial statement section, yielding 39,404 disclosure items preserving original IFRS/note references.
\end{enumerate}

Initial pairing uses the Stage-1 domain-adapted BERT encoder (\texttt{bert-base-cased} continued-pretrained via MLM on SEC EDGAR 10-K MD\&A; \S\ref{sec:embedding-pipeline}, Stage~1) for dense retrieval: each disclosure is encoded into a normalized 768-dimensional vector, and disclosures are matched against regulations via cosine similarity. The top-15 matches per disclosure are retrieved and sampled to the top-5, producing the 20,000 candidate pairs that enter the annotation pipeline. The later TSDAE/SimCSE and supervised contrastive stages cannot be used here because they require labeled regulation--disclosure pairs that do not yet exist at this point in the pipeline.

\section{Prompt Templates}
\label{app:prompts}

The annotation prompt is reproduced  verbatim; the two extraction prompts are released with the code. Placeholders are shown in \{braces\}.

\subsection{IFRS Compliance Annotation Prompt}
\label{app:prompt-compliance}

Each regulation--disclosure pair is presented to all three annotators (GPT-4.1, Claude Sonnet 4, Gemini 2.5 Pro) using the prompt below. The prompt enforces a three-class label with per-class confidence scores returned as a JSON object, parsed via provider-native structured-output enforcement (OpenAI function calling, Anthropic tool use, Gemini controlled generation). Temperature is set to~0 for all providers to ensure deterministic outputs. The user message consists of two fields: \texttt{requirement: \{regulation\_text\}} and \texttt{document: \{disclosure\_text\}}.

\begin{promptbox}{System Prompt: IFRS Compliance Annotation}
You are an expert auditor with perfect knowledge of the IFRS accounting standards. You must assess whether the provided financial statement passage complies with the given regulatory requirement.

\medskip
Assume that only the provided passage is available for evaluation.
If any required disclosure is missing from the passage, it must be considered not complied with.

\medskip
\begin{itemize}
\item Answer ``yes'' if the passage fully satisfies all aspects of the requirement.
\item Answer ``no'' if any part of the requirement is missing, incomplete, or not satisfied in the passage.
\item Answer ``not applicable'' if the requirement is not relevant to the passage.
\end{itemize}

\medskip
In addition to the selected answer, output a confidence score for each possible answer option.

\medskip
\textbf{Output Requirements}
\begin{itemize}
\item Output must be valid JSON only.
\item The confidence score must be a numeric value between 0 and 1 (inclusive).
\item The ``confidence'' field must be a list of three numeric values between 0 and 1 (inclusive).
\item The confidence scores must correspond exactly and in this order to: \texttt{["yes", "no", "not applicable"]}
\item Higher values indicate higher confidence for that option.
\end{itemize}

\medskip
The output must strictly conform to the following JSON format:

\texttt{\{"answer": <"yes"|"no"|"not applicable">, "confidence": [number, number, number]\}}
\end{promptbox}

\section{EM Joint Calibration: Algorithm Details}
\label{app:em-algorithm}

The EM joint calibration procedure alternates between two steps.

\paragraph{Calibration step.}
Each annotator model $m$ has a temperature parameter $T_m$ optimized by minimizing the cross-entropy between $\sigma(\mathbf{p}_{m,i}/T_m)$ and the current soft label estimate $\hat{\mathbf{y}}_i$ over all items~$i$. Minimization uses SciPy \texttt{minimize\_scalar} with the \texttt{bounded} method and bounds $T_m \in [0.1, 20.0]$ (see Appendix~\ref{app:em-audit} for sensitivity analysis).

\paragraph{Label refinement step.}
Soft label estimates are updated as the normalized element-wise product of calibrated model outputs:
\begin{equation}
\hat{\mathbf{y}}_i \;\propto\; \prod_{m} \sigma\!\left(\mathbf{p}_{m,i} / T_m\right).
\end{equation}
The class prior $\boldsymbol{\pi}$ is initialized to $1/K$. On the full 20,000-pair dataset under the adopted bound $T_m \in [0.1, 20.0]$, the procedure converges in 20 iterations (threshold: $\max\Delta < 10^{-5}$): GPT-4.1 fits at $T{=}14.36$ and Gemini~2.5~Pro at $T{=}5.73$, both indicating severe overconfidence in their raw probability outputs; Claude Sonnet~4 converges to $T{=}0.55$, slightly sharpening its already near-calibrated raw outputs.

\section{Per-Model Confusion Matrices}
\label{app:confusion-matrices}

Tables~\ref{tab:cm-openai}--\ref{tab:cm-gemini} present the learned confusion matrices for each LLM annotator, evaluated against the hard-vote consensus on the full 20,000-pair dataset. These matrices reveal systematic prediction biases that motivate the EM calibration procedure (Section~\ref{sec:methodology}).

\begin{table}[H]
\centering
\small
\begin{tabular}{@{}lccc@{}}
\toprule
 & \textbf{Pred: No} & \textbf{Pred: N/A} & \textbf{Pred: Yes} \\
\midrule
\textbf{True: No} & 0.990 & 0.002 & 0.008 \\
\textbf{True: N/A} & 0.472 & 0.527 & 0.001 \\
\textbf{True: Yes} & 0.147 & 0.016 & 0.837 \\
\bottomrule
\end{tabular}
\caption{GPT-4.1 confusion matrix vs.\ hard-vote consensus.}
\label{tab:cm-openai}
\end{table}

\begin{table}[H]
\centering
\small
\begin{tabular}{@{}lccc@{}}
\toprule
 & \textbf{Pred: No} & \textbf{Pred: N/A} & \textbf{Pred: Yes} \\
\midrule
\textbf{True: No} & 0.897 & 0.092 & 0.011 \\
\textbf{True: N/A} & 0.005 & 0.994 & 0.001 \\
\textbf{True: Yes} & 0.114 & 0.002 & 0.884 \\
\bottomrule
\end{tabular}
\caption{Claude Sonnet~4 confusion matrix vs.\ hard-vote consensus.}
\label{tab:cm-anthropic}
\end{table}

\begin{table}[H]
\centering
\small
\begin{tabular}{@{}lccc@{}}
\toprule
 & \textbf{Pred: No} & \textbf{Pred: N/A} & \textbf{Pred: Yes} \\
\midrule
\textbf{True: No} & 0.703 & 0.250 & 0.047 \\
\textbf{True: N/A} & 0.003 & 0.996 & 0.002 \\
\textbf{True: Yes} & 0.060 & 0.016 & 0.925 \\
\bottomrule
\end{tabular}
\caption{Gemini 2.5 Pro confusion matrix vs.\ hard-vote consensus.}
\label{tab:cm-gemini}
\end{table}

\section{Calibration Summary}
\label{app:reliability}

Predicted confidence is compared to observed accuracy for each LLM annotator before and after EM calibration under the adopted bound $T_{\max}{=}20.0$ on the 8,837 human-annotated items. Three observations capture the calibration picture.

\begin{itemize}
\item \textbf{Gemini 2.5 Pro} shows substantial calibration improvement: ECE drops from 0.215 to 0.123 at fitted temperature $T{=}5.73$, with the post-calibration confidence-accuracy curve aligning closely to the diagonal.
\item \textbf{Claude Sonnet 4} is the best-calibrated annotator out of the box (raw ECE 0.211), and EM fits a sharpening temperature $T{=}0.55$ that raises its ECE to 0.276. This is the per-model cost of joint optimization: the temperature that best aligns Anthropic with the consensus soft labels over-sharpens its already near-calibrated raw outputs.
\item \textbf{GPT-4.1} improves materially after calibration (ECE 0.811 to 0.508 at $T{=}14.36$) but remains far from the diagonal. The residual error is structural: GPT-4.1's miscalibration reflects a class-level prediction bias (95.5\% \textsc{Not Complied} on the human-annotated disagreement subset, see Appendix~\ref{app:per-class}) rather than stochastic over- or under-confidence. Temperature scaling cannot correct a label-distribution shift.
\end{itemize}

\section{Human Annotation Platform}
\label{app:platform}

\paragraph{Architecture.}
A purpose-built web application collects independent human judgments. The backend is a FastAPI service with MongoDB for storage; the frontend is a Next.js~16 application. Annotators are assigned items through a priority queue that routes disagreement cases (partial and three-way LLM splits) before unanimous items. Two senior annotators with domain expertise in IFRS reporting oversaw the process, resolving edge cases and ensuring labeling consistency.

\paragraph{Annotation schema.}
Each annotation captures four fields (Table~\ref{tab:annotation-schema}); annotators are blind to LLM predictions:

\begin{table}[h]
\centering
\small
\begin{tabular}{p{2.5cm}p{0.5cm}p{3.5cm}}
\toprule
\textbf{Field} & \textbf{Type} & \textbf{Description} \\
\midrule
\texttt{label}             & enum  & \textsc{Complied}, \textsc{Not Complied}, or \textsc{Not Applicable} \\
\texttt{rationale}         & text  & Free-text justification (1--3 sentences) \\
\texttt{time\_spent}       & float & Auto-tracked annotation duration (seconds) \\
\texttt{flagged\_difficult} & bool & Annotator-flagged ambiguous items \\
\bottomrule
\end{tabular}
\caption{Human annotation schema.}
\label{tab:annotation-schema}
\end{table}

\paragraph{Item selection and routing.}
All 8,795 partial-agreement items (2 of 3 LLMs agree) and all 44 three-way disagreement items are prioritized for human annotation. Unanimous items (11,161) are excluded from human review and reserved for supervised training, so the LLM-unanimous pairs used for embedding fine-tuning are disjoint from the human-labeled disagreement set used for aggregation evaluation.

\section{Data Partition Details}
\label{app:partition}

Table~\ref{tab:partitions} reports the natural partition of the 20,000 LLM-labeled pairs by inter-annotator agreement. The 11,161 unanimous-agreement items (3 of 3 LLMs concur) form the supervised training pool for the Stage~3 contrastive head and never enter the human annotation queue. The 8,795 partial-agreement items (2 of 3 concur) and the 44 three-way disagreement items (all three LLMs disagree) are routed to human annotators in priority order, yielding a routed pool of 8,839 items. The released labels cover 8,837 items (8,794 partial $+$ 43 three-way); the two-item shortfall against the routed pool is one partial-agreement pair and one three-way pair on which the three analysts split three ways and which remained unresolved when the labeling round closed, so both are excluded (\S\ref{sec:human-annotation}). By construction, every item used to train the embedding encoder is excluded from this 8,837-item evaluation set, so the embedding training pool and the classifier/aggregation evaluation set do not overlap.

\begin{table}[t]
\centering
\footnotesize
\setlength{\tabcolsep}{4pt}
\begin{tabular}{@{}lrrl@{}}
\toprule
\textbf{Partition} & \textbf{Count} & \textbf{Labels} & \textbf{Used For} \\
\midrule
Unanimous (3/3)   & 11,161 & 0     & Supervised training \\
Partial (2/3)     & 8,795  & 8,794 & Evaluation \\
Three-way (1/1/1) & 44     & 43    & Evaluation \\
\bottomrule
\end{tabular}
\caption{Data partitions. Items where LLMs agreed unanimously were never sent to humans (0 human labels), keeping the embedding training pool disjoint from the human-labeled evaluation set. The one-label shortfall in each disagreement row (8,795 vs.\ 8,794; 44 vs.\ 43) reflects two routed pairs on which the three analysts split three ways and which remained unresolved when the labeling round closed; they are excluded from the released labels (\S\ref{sec:human-annotation}).}
\label{tab:partitions}
\end{table}

\section{Embedding Pipeline Hyperparameters}
\label{app:hyperparams}

Table~\ref{tab:hyperparams} reports complete hyperparameters for each stage of the three-stage embedding pipeline (Section~\ref{sec:embedding-pipeline}). All stages train on a single RTX~5070 GPU. The longest stage (MLM) completes in approximately 11 hours on the 1.4M-sentence EDGAR-MD\&A corpus, while TSDAE, SimCSE, and supervised contrastive each complete in under 90 minutes on the 42,893-sentence adaptation set and 4,930-pair supervised set respectively. The cased tokenizer is retained to preserve capitalization-sensitive acronyms (IFRS, IAS, GAAP) throughout the pipeline.

\begin{table*}[t]
\centering
\small
\begin{tabular}{@{}lllll@{}}
\toprule
\textbf{Parameter} & \textbf{Stage 0 (MLM)} & \textbf{Stage 1A (TSDAE)} & \textbf{Stage 1B (SimCSE)} & \textbf{Stage 2 (Supervised)} \\
\midrule
Base model & bert-base-cased & Stage 0 output & Stage 0 output & Stage 1 output \\
Epochs & 3 & 1 & 1 & 4 \\
Batch size & 4 (accum.\ 4) & 8 & 128 & 16 \\
Learning rate & 5e-5 & 3e-5 & 3e-5 & 2e-5 \\
fp16 & Yes & No & Yes & Yes \\
Warmup & --- & 10\% & 10\% & 10\% \\
Temperature ($\tau$) & --- & --- & 0.05 & 0.05 \\
Token deletion & 15\% (MLM) & 60\% (TSDAE) & --- & --- \\
Tied weights & --- & Yes & --- & --- \\
Grad.\ checkpoint & No & No & Yes & No \\
Training data & EDGAR 10-K & 42,893 sentences & 42,893 sentences & 4,930 pos.\ pairs \\
Loss & MLM & Denoising AE & MNRL & MNRL \\
\bottomrule
\end{tabular}
\caption{Hyperparameters for each stage of the embedding pipeline. MNRL = MultipleNegativesRankingLoss. EDGAR 10-K sentences are filtered to 150--400 characters from Section~7 (MD\&A) of annual filings (1993--2020).}
\label{tab:hyperparams}
\end{table*}

\section{Extended Retrieval Results}
\label{app:retrieval}

Tables~\ref{tab:retrieval-seen} and~\ref{tab:retrieval-unseen} report retrieval performance across all five embedding checkpoints on seen and unseen regulations respectively. Three observations stand out. First, SimCSE~+~Supervised is the strongest model on unseen regulations across every metric (R@15~$=$~45.3, MAP@15~$=$~0.125, NDCG@15~$=$~0.188), making it the preferred encoder for zero-shot retrieval generalization. Second, plain SimCSE without supervised fine-tuning under-performs the MLM-only baseline on seen regulations (R@15~$=$~15.8 vs.\ 20.9), indicating that contrastive dropout-noise alone does not generalize the encoder beyond its MLM initialization when training on a small in-domain corpus. Third, TSDAE~+~Supervised wins on R@5 (both seen and unseen), suggesting that denoising auto-encoding produces more discriminative top-ranked candidates while SimCSE produces a flatter but more recall-favorable ranking.

\begin{table*}[t]
\centering
\small
\begin{tabular}{@{}lrrrrrrr@{}}
\toprule
\textbf{Model} & \textbf{R@1} & \textbf{R@5} & \textbf{R@10} & \textbf{R@15} & \textbf{MRR} & \textbf{MAP@15} & \textbf{NDCG@15} \\
\midrule
MLM only & 2.5 & 8.9 & 17.0 & 20.9 & 0.060 & 0.034 & 0.059 \\
+ TSDAE & 2.1 & 7.7 & 14.5 & 19.8 & 0.053 & 0.034 & 0.058 \\
+ SimCSE & 1.5 & 7.4 & 11.3 & 15.8 & 0.046 & 0.026 & 0.046 \\
+ TSDAE + Sup & 2.5 & \textbf{11.9} & 22.1 & \textbf{27.5} & \textbf{0.080} & \textbf{0.050} & \textbf{0.084} \\
+ SimCSE + Sup & 2.5 & 11.3 & 23.2 & 26.4 & 0.073 & 0.042 & 0.077 \\
\bottomrule
\end{tabular}
\caption{Seen-regulation retrieval metrics.}
\label{tab:retrieval-seen}
\end{table*}

\begin{table*}[t]
\centering
\small
\begin{tabular}{@{}lrrrrrrr@{}}
\toprule
\textbf{Model} & \textbf{R@1} & \textbf{R@5} & \textbf{R@10} & \textbf{R@15} & \textbf{MRR} & \textbf{MAP@15} & \textbf{NDCG@15} \\
\midrule
MLM only & 3.8 & 11.3 & 21.7 & 28.3 & 0.086 & 0.076 & 0.118 \\
+ TSDAE & 0.9 & 12.3 & 22.6 & 26.4 & 0.069 & 0.064 & 0.106 \\
+ SimCSE & 0.9 & 12.3 & 15.1 & 27.4 & 0.078 & 0.070 & 0.110 \\
+ TSDAE + Sup & 4.7 & 18.9 & 32.1 & 41.5 & 0.122 & 0.110 & 0.171 \\
+ SimCSE + Sup & 3.8 & \textbf{20.8} & \textbf{33.0} & \textbf{45.3} & \textbf{0.137} & \textbf{0.125} & \textbf{0.188} \\
\bottomrule
\end{tabular}
\caption{Unseen-regulation retrieval metrics. SimCSE + Supervised achieves the strongest zero-shot generalization.}
\label{tab:retrieval-unseen}
\end{table*}

\section{Label Source Ablation: Extended Results}
\label{app:ablation}

Training stage 3 on 4,930 LLM-unanimous pairs yields overlapping 95\% bootstrap confidence intervals with training on 1,616 human-annotated pairs (Table~\ref{tab:label-ablation}). Since the LLM pairs are a zero-cost byproduct of the annotation pipeline, this supports using them as a practical alternative to expert labeling in stage 3.

% Label source ablation table
\begin{table}[H]
\centering
\footnotesize
\setlength{\tabcolsep}{4pt} % tighter column spacing

\begin{tabular}{llcc}
\toprule
\textbf{Model} & \textbf{Labels} & \textbf{Seen@5} & \textbf{Unseen@15} \\
\midrule
TSDAE+Sup & Human (1.6k) & 6.2 & 36.0 \\
TSDAE+Sup & LLM (4.9k)   & 4.7 & 40.0 \\
SimCSE+Sup & Human (1.6k) & 7.8 & 32.0 \\
SimCSE+Sup & LLM (4.9k)   & 5.7 & 48.0 \\
\bottomrule
\end{tabular}

\caption{Label source results on shared test set (1,326 items). All CIs overlap between LLM and Human rows.}
\label{tab:label-ablation}
\end{table}

Table~\ref{tab:ablation} isolates the effect of supervised label source on the final retrieval encoder. The test set is held fixed (1,326 items, 192 seen regulations, 25 unseen regulations) while the Stage~3 supervised contrastive head is retrained on two label sources: (i) 1,616 human-validated positive pairs from the 8,837 human-annotated items, and (ii) 4,930 LLM-unanimous positive pairs harvested from the 11,161 unanimous-agreement subset.

Two findings stand out. First, neither label source dominates: human labels score higher on seen regulations, while LLM-unanimous labels score higher on unseen R@15 for both backbones (48.0 vs.\ 32.0 for SimCSE~+~Supervised; 40.0 vs.\ 36.0 for TSDAE~+~Supervised). Second, the gap is largest precisely where data scarcity hurts most: on unseen regulations, where the model must generalize beyond its training distribution. The 3.05$\times$ larger positive-pair budget from LLM unanimity dominates whatever label-noise penalty unanimous LLM agreement carries. Because LLM unanimity is a zero-cost byproduct of the annotation pipeline while human annotation requires paid labor, this result supports weak supervision as a practical alternative to expert labeling in this domain when three diverse LLM judges agree.

\begin{table}[t]
\centering
\footnotesize
\setlength{\tabcolsep}{5pt}
\begin{tabular}{@{}llcc@{}}
\toprule
& & \textbf{Seen} & \textbf{Unseen} \\
\textbf{Model} & \textbf{Labels (n)} & \textbf{R@5\,/\,R@15} & \textbf{R@5\,/\,R@15} \\
\midrule
TSDAE+S  & Human (1.6k) & 6.2\,/\,14.6 & 8.0\,/\,36.0  \\
TSDAE+S  & LLM (4.9k)   & 4.7\,/\,12.0 & 16.0\,/\,40.0 \\
SimCSE+S & Human (1.6k) & 7.8\,/\,15.6 & 20.0\,/\,32.0 \\
SimCSE+S & LLM (4.9k)   & 5.7\,/\,13.5 & 12.0\,/\,48.0 \\
\bottomrule
\end{tabular}
\caption{Label-source ablation on a fixed test set (1,326 items, 192 seen regs, 25 unseen regs). Each cell reports R@5\,/\,R@15 (\%). 95\% bootstrap CIs omitted for space.}
\label{tab:ablation}
\end{table}

\section{Per-Class Detailed Metrics}
\label{app:per-class}

Table~\ref{tab:per-class} decomposes each hard predictor's performance by the three target classes. Three failure patterns emerge. (1) OpenAI's ``yes'' row is near empty: only 233 of 8,837 predictions are positive, yielding undefined precision and recall under the macro definition. Its ``no'' prediction is high-recall (0.98) but low-precision (0.16), confirming the 95.5\% over-prediction of \textsc{Not Complied} reported in Section~\ref{sec:aggregation-results}. (2) Anthropic is balanced across classes but uniformly low on the minority ``yes'' class (F1~$=$~0.20). (3) Gemini achieves the strongest per-class F1 on ``yes'' (0.59) and ``not applicable'' (0.85) but degrades on ``no'' (0.16), mirroring the inverse of OpenAI's bias. Hard vote partially compensates by averaging the biases (yes F1~$=$~0.42, no F1~$=$~0.32, n/a F1~$=$~0.78), but no hard predictor exceeds the learned ensemble classifier (Section~\ref{sec:learned-en-classifier}) on any class. These class-specific failure modes are invisible at the aggregate W-F1 level and motivate the per-class oracle ceiling reported in Table~\ref{tab:agg-extended}.

\paragraph{Hardest standards.} Per-standard scores use per-item out-of-fold predictions of the reported classifier configuration, re-run to save them (W-F1 0.813; over five XGBoost seeds 0.814, s.d.\ 0.001). This re-run does not reproduce the reported 0.817 exactly. Weighted F1 is lowest on IFRS~6 (0.735), IFRS~3 (0.772), IFRS~13 (0.776), IFRS~5 (0.779) and IFRS~2 (0.780). \textsc{Not Complied} is the weakest class throughout; among standards with at least 50 \textsc{Not Complied} items its F1 is lowest on IFRS~3 (0.42), IFRS~2 (0.44), IFRS~16 (0.46) and IFRS~5 (0.46). The rarest class is not the hardest: \textsc{Complied}, 10.3\% of the human labels, reaches F1 0.71 against 0.50 for \textsc{Not Complied}.

\begin{table}[t]
\centering
\small
\begin{tabular}{@{}llrrr@{}}
\toprule
\textbf{Predictor} & \textbf{Class} & \textbf{Prec.} & \textbf{Rec.} & \textbf{F1} \\
\midrule
OpenAI & yes & --- & --- & --- \\
OpenAI & no & 0.16 & 0.98 & 0.28 \\
OpenAI & not applicable & 0.80 & 0.02 & 0.04 \\
\midrule
Anthropic & yes & 0.14 & 0.35 & 0.20 \\
Anthropic & no & 0.29 & 0.27 & 0.28 \\
Anthropic & not applicable & 0.83 & 0.82 & 0.83 \\
\midrule
Gemini & yes & 0.73 & 0.49 & 0.59 \\
Gemini & no & 0.21 & 0.13 & 0.16 \\
Gemini & not applicable & 0.80 & 0.90 & 0.85 \\
\midrule
Hard Vote & yes & 0.79 & 0.28 & 0.42 \\
Hard Vote & no & 0.23 & 0.50 & 0.32 \\
Hard Vote & not applicable & 0.85 & 0.72 & 0.78 \\
\bottomrule
\end{tabular}
\caption{Per-class precision, recall, and F1 for all predictors on 8,837 human-annotated items. OpenAI ``yes'' metrics are near-zero (only 233 of 8,837 predictions are ``yes'').}
\label{tab:per-class}
\end{table}

\section{Aggregation Methods: Extended Benchmark}
\label{app:aggregation-extended}

Table~\ref{tab:agg-extended} reports the full classification benchmark on 8,837 human-annotated items, ranked by W-F1. The table extends Section~\ref{sec:aggregation-results} (Table~\ref{tab:classification}) with macro-F1 and Brier-score columns, and adds the per-item oracle (theoretical ceiling) and the learned ensemble classifier (Section~\ref{sec:learned-en-classifier}) for direct comparison. The 11 aggregation strategies span majority voting (Hard/Soft Vote), confusion-matrix EM (Dawid--Skene, One-Coin DS), competence estimation (MACE, M-MSR, GLAD, Wawa), classifier-augmented consensus (CROWDLAB), and temperature-scaled EM under both bound choices (Section~\ref{sec:eval-protocol}). Dashes indicate metrics not produced by the method (e.g., MACE, CROWDLAB, and One-Coin DS output discrete labels rather than calibrated class probabilities, so ECE and Brier are undefined).

\begin{table*}[t]
\centering
\small
\begin{tabular}{lcccccc}
\toprule
\textbf{Strategy} & \textbf{Acc} & \textbf{W-F1} & \textbf{M-F1} & \textbf{$\kappa$} & \textbf{ECE} & \textbf{Brier} \\
\midrule
Oracle (ceiling)              & 0.952 & 0.949 & 0.900 & 0.882 & 0.031 & 0.102 \\
\midrule
\textbf{Learned Ensemble (proposed)} & \textbf{0.824} & \textbf{0.817} & --    & \textbf{0.556} & \textbf{0.068} & --    \\
\midrule
MACE \cite{hovy_learning_2013}                   & 0.748 & 0.722 & 0.561 & 0.312 & --    & --    \\
CROWDLAB \cite{goh_crowdlab_2022}                & 0.791 & 0.720 & --    & 0.340 & --    & --    \\
Gemini single \cite{google_gemini25_2025}   & 0.737 & 0.713 & 0.532 & 0.278 & 0.215 & 0.478 \\
M-MSR \cite{ma_mmsr_2020}                        & 0.672 & 0.687 & --    & 0.256 & --    & --    \\
Anthropic single \cite{anthropic_claude4_2025}   & 0.672 & 0.687 & 0.499 & 0.256 & 0.211 & 0.579 \\
One-Coin DS \cite{zhang_spectral_em_2016}        & 0.668 & 0.679 & --    & 0.238 & --    & --    \\
Soft Vote                                        & 0.642 & 0.672 & 0.507 & 0.255 & 0.290 & 0.538 \\
EM Calibrated ($T_{\max}{=}20$) \cite{guo_calibration_2017} & 0.647 & 0.672 & --    & 0.245 & 0.292 & --    \\
Hard Vote                                        & 0.641 & 0.671 & 0.505 & 0.254 & --    & --    \\
Dawid--Skene \cite{dawid_maximum_1979}           & 0.641 & 0.671 & 0.504 & 0.254 & --    & --    \\
Wawa \cite{ustalov_crowdkit_2024}                & 0.642 & 0.671 & --    & 0.254 & --    & --    \\
GLAD \cite{whitehill_whose_2009}                 & 0.642 & 0.671 & --    & 0.254 & --    & --    \\
EM Calibrated ($T_{\max}{=}5$, default)          & 0.456 & 0.503 & 0.381 & 0.149 & 0.461 & 0.980 \\
GPT-4.1 single \cite{openai_gpt41_2025}           & 0.183 & 0.095 & 0.198 & 0.019 & 0.811 & 1.621 \\
\bottomrule
\end{tabular}
\caption{Classification benchmark on 8,837 human-annotated items, ranked by W-F1. Bold row: the learned ensemble classifier (Section~\ref{sec:learned-en-classifier}). EM Calibrated under the textbook $T_{\max}{=}5$ bound is included to make the 17-point bound penalty visible (Appendix~\ref{app:em-audit}).}
\label{tab:agg-extended}
\end{table*}

\section{Semi-Supervised EM}
\label{app:semi-sup-em}

This appendix tests whether clamping human gold labels during the EM label-refinement step improves calibration. Gold items are held fixed during the E-step; the calibration step optimizes temperatures using a mixture of inferred and known targets. Across gold budgets of 100, 500, and 1,000 items, W-F1 stays at 0.672--0.673 (indistinguishable from unsupervised EM) and ECE remains at 0.290--0.293. At budgets of 2,000 and 5,000, performance collapses: W-F1 drops to 0.582 and 0.513 respectively, with ECE rising to 0.322 and 0.364. Anthropic's fitted temperature drifts from a well-calibrated $T{=}0.55$ at zero budget upward to $T{=}1.45$ at budget 2,000 and $T{=}4.69$ at budget 5,000.

\paragraph{Root cause.}
All 8,837 gold labels are on LLM-disagreement items, the hardest cases. At small budgets, clamped items are a negligible fraction of the M-step loss. At budgets $\geq 2,000$, clamped items dominate and the M-step is forced to find temperatures consistent with hard-case ground truth, which pushes Anthropic's well-calibrated temperature toward over-flattening. Representative random sampling across all agreement tiers would be required for gold anchoring to help. The main text therefore keeps the unsupervised-only EM.

\section{Classifier Ablations}
\label{app:router-ablations}

\paragraph{Architecture comparison (probability-only features, 29-d).}
Four classifier architectures are compared over the 29-dimensional probability feature vector: confidence-weighted selection, stacked generalization, direct 3-class XGBoost, and 4-class model selection. All four saturated within a 0.003 W-F1 band of 0.727 under 5-fold cross-validation on the 8,837-item set, confirming that the probability features alone carry limited routing signal regardless of architecture choice.

\paragraph{Embedding encoder comparison (hybrid features).}
Adding 1,536-d frozen SBERT embeddings to the 29-d probability features lifts performance by 8 points (from 0.727 to 0.808 W-F1). Within the family of compatible encoders tested (TSDAE, BERT-SEC MLM, TSDAE + Supervised, and vanilla \texttt{bert-base-cased}), all hybrid variants fell within a 0.4-point W-F1 band, suggesting that at this scale the choice of base encoder is dominated by the contribution of the embedding modality itself rather than the specific adaptation strategy. Notably, TSDAE-unsupervised (0.808) edged out TSDAE+Supervised (0.804), suggesting that the supervised contrastive objective optimizes for retrieval similarity rather than for the discriminative features the classifier requires; the TSDAE Stage-2A (unsupervised) checkpoint is therefore the encoder used as the frozen feature extractor in the headline learned-ensemble result. The full 5-checkpoint encoder sweep including the SimCSE family is left to follow-up work.

\paragraph{Optimization.}
Optuna hyperparameter search (TPE, 100 trials) on the XGBoost head, targeting the \textsc{Not Complied} bottleneck identified in Appendix~\ref{app:per-class}, achieved 0.817 W-F1, the headline classification result. Class-weighting alone (without hyperparameter search) reaches 0.808. Augmenting training with the 11,161 LLM-unanimous items as silver labels does not help (0.810). Full search trajectories will be included in the code release.

\paragraph{Agreement with unanimous LLM labels.}
Trained only on disagreement items, the classifier agrees with the LLM labels on the 11,161 unanimous items at 0.624 W-F1 (accuracy 0.663). On these items the agreement features take values absent from training, so the score reflects distribution shift as much as label quality; with no human labels on this subset, it measures agreement with the LLM label, not accuracy.

\paragraph{Fine-tuned BERT baseline (rejected).}
End-to-end fine-tuning of the domain-adapted BERT (3 epochs, batch 16) yields 0.755 W-F1, below the frozen hybrid at 0.808; with ${\sim}7$K training items per fold, frozen embeddings plus gradient boosting is more data-efficient.

\section{EM Temperature Bound Sensitivity}
\label{app:em-audit}

The choice of upper bound for the per-model temperature parameter is the single largest non-data factor in EM calibration accuracy. Table~\ref{tab:em-tbound} reports W-F1 and ECE under three bounds. The conventional textbook bound of $T_{\max}{=}5.0$, used in most temperature-scaling literature, is binding for both OpenAI ($T{=}5.0$ at bound) and Gemini ($T{=}5.0$ at bound). Forcing two of three annotators against the bound suppresses the differential reweighting that EM relies on, collapsing W-F1 to 0.503, well below soft vote (0.672) and below every individual aggregator in Table~\ref{tab:agg-extended}. Relaxing the bound to $T_{\max}{=}20.0$ lets OpenAI settle at $T{=}14.36$ and Gemini at $T{=}5.73$, recovering W-F1 to 0.672 (matching soft vote) and reducing ECE from 0.461 to 0.292. The bound $[0.1, 20.0]$ is adopted throughout the main results (Section~\ref{sec:methodology}, Appendix~\ref{app:em-algorithm}). The 17-point W-F1 gap between the two bound choices is the silent EM temperature-bound error documented as a diagnostic of multi-LLM aggregation in this work.

\begin{table}[t]
\centering
\footnotesize
\setlength{\tabcolsep}{4pt}
\begin{tabular}{@{}cccccc@{}}
\toprule
\textbf{$T_{\max}$} & \textbf{$T_{\text{oai}}$} & \textbf{$T_{\text{ant}}$} & \textbf{$T_{\text{gem}}$} & \textbf{W-F1} & \textbf{ECE} \\
\midrule
5.0  & 5.00$^{\star}$  & 0.95 & 5.00$^{\star}$ & 0.503 & 0.461 \\
10.0 & 10.00$^{\star}$ & 0.66 & 5.84           & 0.642 & 0.301 \\
20.0 & 14.36           & 0.55 & 5.73           & 0.672 & 0.292 \\
\bottomrule
\end{tabular}
\caption{Temperature bound sensitivity. $^{\star}$ marks fitted temperatures pinned at the upper bound. The $[0.1,5.0]$ bound introduces a 17-point W-F1 penalty by constraining two of three models to the bound.}
\label{tab:em-tbound}
\end{table}

\end{document}